\documentclass[conference,a4paper]{IEEEtran}
\IEEEoverridecommandlockouts

\usepackage{cite}
\usepackage{amsmath,amssymb,amsfonts}
\usepackage{graphicx}
\usepackage{textcomp}
\usepackage[T1]{fontenc}
\usepackage{booktabs}
\usepackage{xcolor}
\usepackage[bookmarks=false]{hyperref}
\usepackage{balance}

\hypersetup{
    colorlinks=true,
    linkcolor=black,
    urlcolor=blue,
    citecolor=black
}

\def\BibTeX{{\rm B\kern-.05em{\sc i\kern-.025em b}
\kern-.08emT\kern-.1667em\lower.7ex\hbox{E}\kern-.125emX}}

\title{When Clean Signals Are Not Enough: Detecting Structural 
Ambiguity for Safe Wearable Stress Classification}

\author{
\IEEEauthorblockN{Saba A. Farahani}
\IEEEauthorblockA{
University of California, Irvine\\
Irvine, CA, USA\\
fazizaba@uci.edu}
\and
\IEEEauthorblockN{Hung Cao}
\IEEEauthorblockA{
University of California, Irvine\\
Irvine, CA, USA\\
hungcao@uci.edu}
\and
\IEEEauthorblockN{Amir M. Rahmani}
\IEEEauthorblockA{
University of California, Irvine\\
Irvine, CA, USA\\
a.rahmani@uci.edu}
}

\begin{document}

\maketitle

\begin{abstract}
Wearable stress classifiers can achieve strong average performance while
failing completely for a particular individual. On WESAD, a Random Forest
reaches 93.0\% mean accuracy yet yields $F1=0$ for Subject~14, whose
cross-signal coupling weakens near stress onset. We call this
\emph{structural ambiguity}: individually plausible physiological channels
form an inter-signal pattern that is poorly supported by the person's
non-stress reference. We introduce the Individual Conformal Coupling
Monitor (ICCM), a lightweight and transparent pre-inference monitor that
quantifies subject-specific coupling divergence and routes each window to
classify, defer, or abstain without retraining the downstream classifier.
Across WESAD ($N=15$) and Stress-Predict ($N=35$), full-cohort Pearson
associations between ambiguity and accuracy are negative ($r=-0.607$,
$p=0.016$; $r=-0.412$, $p=0.014$). Robustness analyses temper this finding:
rank correlations are not significant, and the WESAD association
disappears when Subject~14 is removed. ICCM changes false-positive counts
from 29 to 27 and 94 to 92, although neither paired change is significant.
It withholds 3 of Subject~14's 21 stress windows but does not repair the
missed-stress failure. These results position ICCM as an interpretable
signal of unsupported physiology and individual failure, rather than a
stand-alone safety guarantee.
\end{abstract}


\noindent\textbf{Code Availability.}
Code is available on 
\href{https://github.com/Saba-Farahani/structural-ambiguity-iccm.git}
{GitHub}.

\section{Introduction}

\begin{figure}[t]
    \centering
    \includegraphics[width=\columnwidth]{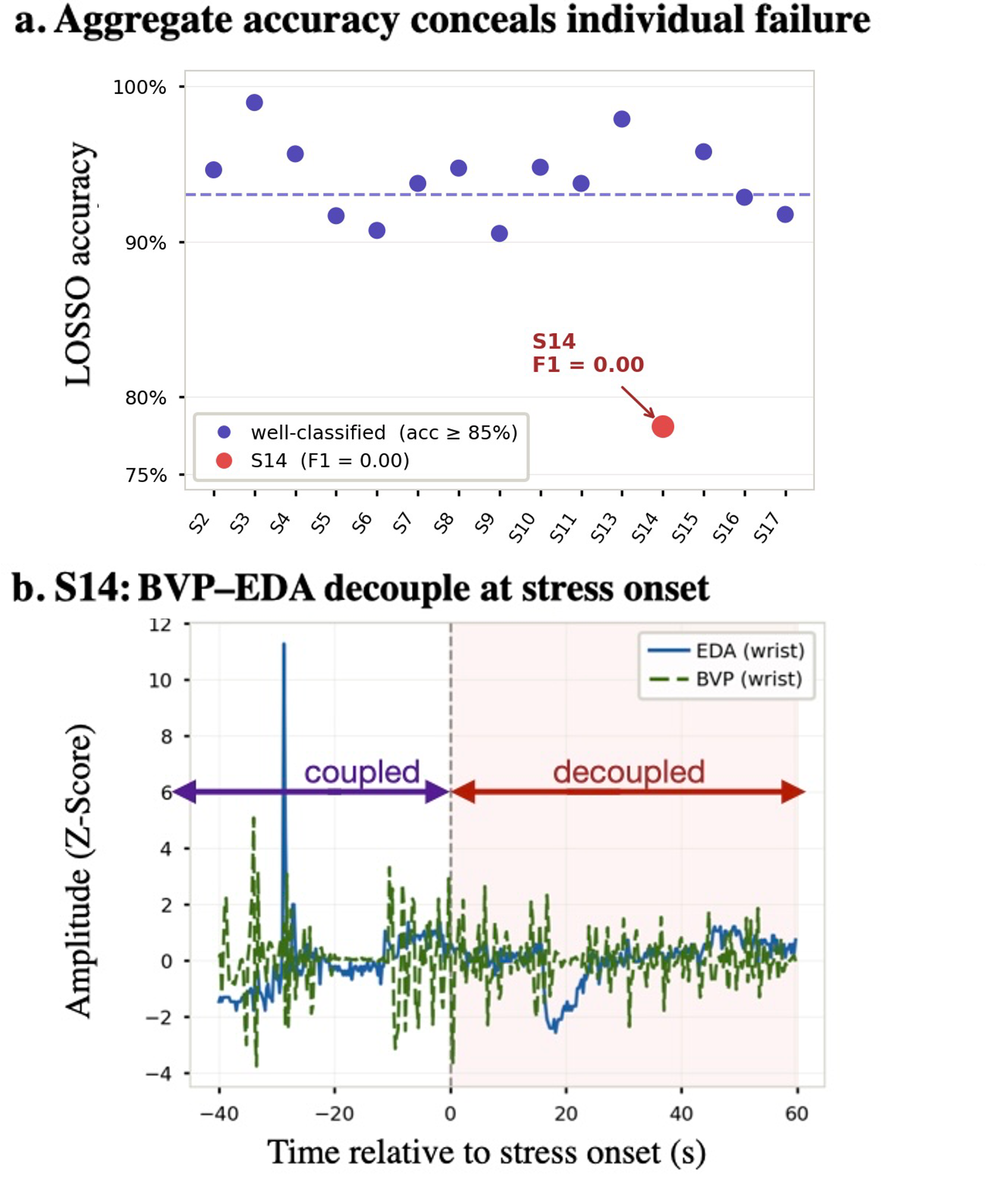}
    \caption{Aggregate accuracy conceals Subject~14's
    missed-stress failure ($F1=0$). EDA and BVP decouple
    near stress onset; residual artifacts cannot be excluded.}
    \label{fig:motivation}
\end{figure}

\begin{figure*}[!t]
    \centering
    \includegraphics[width=\textwidth]{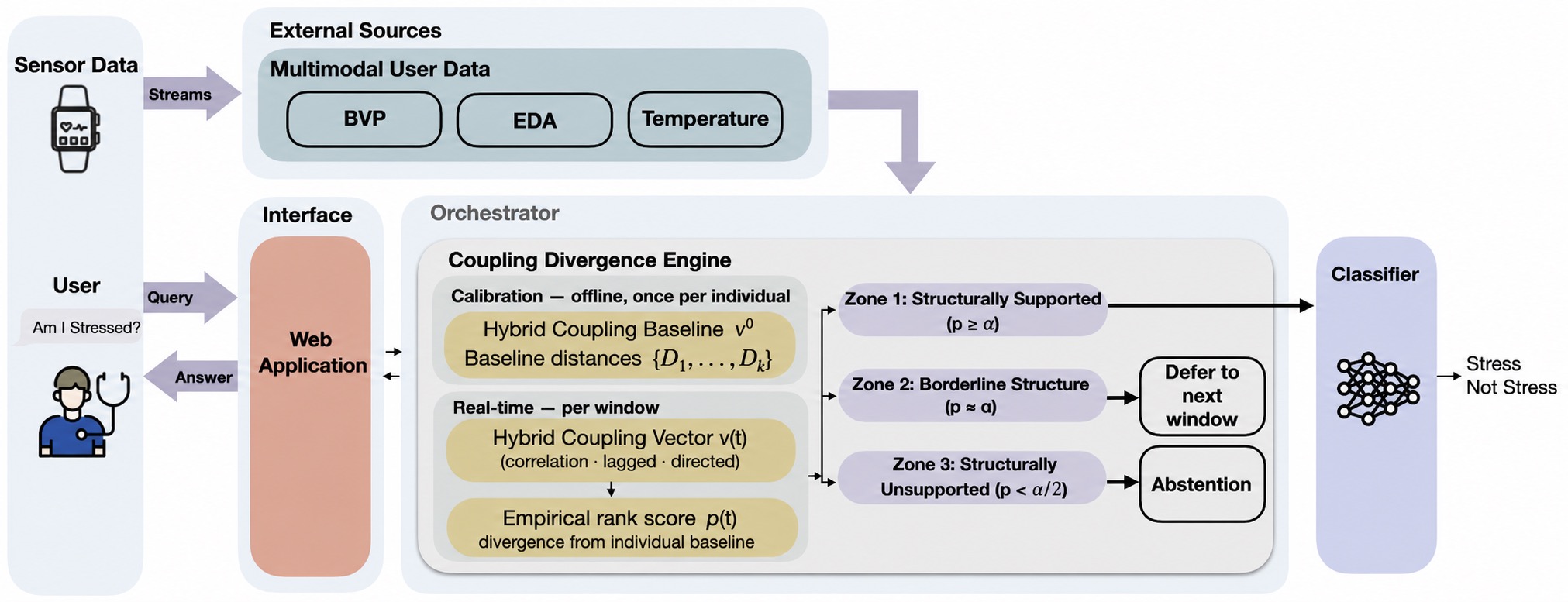}
    \caption{ICCM system architecture.
    Wearable sensor streams (BVP, EDA, TEMP) are passed
    to the Coupling Divergence Engine within the
    Orchestrator. During offline calibration, the engine
    computes a subject-specific hybrid coupling baseline
    $\mathbf{v}^0$ and baseline distances
    $\{D_1,\ldots,D_k\}$ from resting-state windows. At
    inference time, each 60-second window is evaluated
    using a hybrid coupling vector $\mathbf{v}(t)$
    combining Pearson correlation, max-lag
    cross-correlation, and Granger-style directed coupling,
    converted to an empirical conformal-style rank score $p(t)$, and routed
    through a 3-Zone Safety Gate: Zone~1 ($p \geq \alpha$)
    passes to the classifier, Zone~2 ($p \approx \alpha$)
    defers to the next window, and Zone~3 ($p < \alpha/2$)
    triggers abstention. The term ``safe'' denotes the system objective,
    not a clinical guarantee.}
    \label{fig:architecture}
\end{figure*}

Wearable physiological classifiers for stress detection
often report high mean leave-one-subject-out (LOSO)
accuracy while concealing severe failures for specific
individuals~\cite{schmidt2018wesad,gjoreski2017monitoring}.
On WESAD~\cite{schmidt2018wesad}, a Random Forest achieves
93\% mean accuracy yet yields $F1=0.000$ for Subject~14,
whose EDA--BVP coupling weakens near stress onset. The
channels contain no missing samples, but a diagnostic screen
cannot establish that they are artifact-free. We use
\emph{structural ambiguity} for the operational condition in
which the observed inter-signal relationship is poorly
supported by a subject-specific reference.

In real-world wearable health systems, such failures carry
direct clinical consequences: false stress alerts can
trigger unnecessary interventions, contribute to alarm
fatigue, and erode patient and clinician trust in
physiological monitoring---barriers increasingly recognized
as central obstacles to wearable AI adoption in
healthcare~\cite{sendelbach2013alarm}.

Existing approaches address wearable classifier failures
through improved architectures, data augmentation, or
distributionally robust optimization~\cite{sagawa2020dro}.
These methods can improve average accuracy, but they do not
answer a pre-inference safety question: \emph{is the current
physiological coupling structure supported by this
individual's baseline?} Confidence scores and output-level uncertainty estimates are
computed after the classifier has processed the input and do
not directly detect structurally invalid inputs before
inference occurs.

We introduce the Individual Conformal Coupling Monitor
(ICCM), which calibrates a subject-specific non-stress
coupling reference and applies a three-zone gate to classify,
defer, or abstain before inference. ICCM requires no model
retraining and is classifier-external in implementation
(Fig.~\ref{fig:architecture}); multi-architecture performance
remains untested.

This paper makes three contributions:
\begin{itemize}
    \item We define \emph{structural ambiguity} as
    insufficiently supported inter-signal coupling despite
    individually plausible channels.
    \item We introduce ICCM, a subject-specific,
    classifier-external three-zone routing monitor.
    \item We evaluate ICCM on WESAD ($N=15$) and
    Stress-Predict ($N=35$), including robustness and
    selective-outcome analyses.
\end{itemize}

\section{Related Work}
\textbf{Wearable Stress Detection.}
Multimodal wearable stress detection has been widely studied
using BVP, EDA, and skin temperature, with Random Forests
and other models achieving high average LOSO performance on
datasets such as
WESAD~\cite{schmidt2018wesad,gjoreski2017monitoring,can2019stress}.
ICCM instead monitors whether a window is supported by a
subject-specific non-stress reference; its Stress-Predict
extension also uses stress labels from LOSO training subjects.

\textbf{Signal Quality and Abstention.}
Signal-quality methods detect hardware degradation, motion
artifacts, or poor recordings~\cite{orphanidou2015signal},
whereas ICCM checks inter-signal coupling. Behavior-adaptive
models also show interpretable coupling changes across
behavioral phases~\cite{asadi2025bace}. Selective prediction
supports abstention under high risk~\cite{chow1957optimum,geifman2017selective},
and clinical AI uses conformal and Bayesian uncertainty for
abstention~\cite{angelopoulos2022gentle}. ICCM provides a
physiological, classifier-external routing reason, but its
overlapping baseline and rank windows preclude a formal
conformal-coverage guarantee here.
\section{Method}
ICCM is a deterministic physiological filter using hybrid
coupling nonconformity and empirical rank calibration. It has
three components (Fig.~\ref{fig:architecture}).

\subsection{External Sources}
External Sources provide Empatica E4 BVP (64~Hz), EDA
(4~Hz), and TEMP (4~Hz). HR is derived from BVP by sliding
5-s peak detection; signals use 60-s windows with a 30-s step.

\subsection{Interface}
The Interface returns the classifier output when structurally
supported, or reports insufficient physiological evidence.

\subsection{Orchestrator}
The Orchestrator performs coupling analysis and routing in two phases.

\textbf{Phase 1: Calibration.}
The calibration phase runs once per individual on known
non-stress windows. For each window $w_k$, a hybrid
coupling vector is computed over signal pairs
$(x,y) \in \{$EDA--HR, EDA--TEMP, HR--TEMP$\}$:
\begin{equation}
\begin{split}
    \mathbf{v}(w_k) = [
        &\rho_{EH},\; \rho_{ET},\; \rho_{HT}, \\
        &\ell_{EH},\; \ell_{ET},\; \ell_{HT}, \\
        &g_{E{\to}H},\; g_{H{\to}E},\; g_{T{\to}H}]
\end{split}
\end{equation}
\noindent where $\rho_{xy}$ is the absolute Pearson
correlation; $\ell_{xy}$ is the maximum absolute
cross-correlation over physiological delays
$\tau \in [1\text{s},10\text{s}]$; and
$g_{x \to y} = \min(-\log p_{\mathrm{GC}}, 10)/10$ is
a Granger-style directed coupling score normalized to
$[0,1]$. The subject-specific baseline and calibration
distances are:
\begin{equation}
    \mathbf{v}^0 = \frac{1}{K}\sum_{k=1}^{K}
    \mathbf{v}(w_k), \quad
    D_k = \|\mathbf{v}(w_k)-\mathbf{v}^0\|_2
\end{equation}
\noindent The set $\mathcal{D}=\{D_1,\ldots,D_K\}$
defines this individual's normal coupling variation.

\textbf{Phase 2: Real-time Monitoring.}
At each window $t$, the Orchestrator computes
$\mathbf{v}(t)$ and evaluates an empirical conformal-style rank score:
\begin{equation}
    p(t) = \frac{1+|\{k:D_k \geq D(t)\}|}{K+1},
    \quad D(t) = \|\mathbf{v}(t)-\mathbf{v}^0\|_2
\end{equation}
\noindent A low $p(t)$ indicates that the current
coupling deviates from this individual's baseline more
than most calibration windows, providing evidence of
structural ambiguity.
The same overlapping windows estimate $\mathbf{v}^0$ and
$\mathcal{D}$, so exchangeability and split-conformal
independence are not established. At $\alpha=0.05$, Zone~3
is reachable only for $K\ge40$; 60-s windows with a 30-s
step require at least 20.5 minutes of contiguous calibration.
For WESAD, $K=72$--76 because all labeled non-stress periods
(baseline, amusement, and meditation), not rest alone, are used.

\subsection{Output: 3-Zone Safety Gate}
The Orchestrator routes each window based on $p(t)$
with $\alpha=0.05$:
\begin{itemize}
    \item \textbf{Zone~1} ($p(t) \geq \alpha$):
    structurally supported. Forwarded to the downstream
    Random Forest classifier~\cite{breiman2001random}.
    \item \textbf{Zone~2} ($\alpha/2 \leq p(t) < \alpha$):
    borderline. Window is withheld from classification; no prediction is issued.
    \item \textbf{Zone~3} ($p(t) < \alpha/2$):
    structurally unsupported. Abstention is triggered; this
    routing action is not itself a clinical safety guarantee.
\end{itemize}

\subsection{Protocol-Aware Coupling Selection}
In single-protocol datasets (WESAD), magnitude-based
divergence from individual baseline is sufficient to
detect coupling collapse. We use six features ($\rho$
and $\ell$ only), omitting directed coupling, which adds
noise when protocol variability is low. In multi-protocol
datasets (Stress-Predict), we use all nine features
($\rho$, $\ell$, and $g$) with a direction-aware score:
\begin{equation}
    D_{\mathrm{dir}}(t) =
    1 - \cos\!\left(\Delta\mathbf{v}(t),\,
    \boldsymbol{\mu}_{\Delta}\right)
\end{equation}
\noindent where $\Delta\mathbf{v}(t)=
\mathbf{v}(t)-\mathbf{v}^0$ and
$\boldsymbol{\mu}_{\Delta}$ is the population mean
coupling-change direction estimated from training
subjects under LOSO. Both $\boldsymbol{\mu}_{\Delta}$ and
the empirical routing distribution use \emph{labeled stress
windows} from LOSO training subjects; only the test subject's
reference is label-free. The two dataset-specific
configurations were selected after ablation and remain
exploratory. A fixed unsupervised configuration does not
transfer to Stress-Predict ($r=0.474$), while a fixed
direction-aware configuration does not transfer to WESAD
($r=0.209$).

\subsection{Signal-Quality Diagnostic}
For Subject~14, we screened finiteness, channel ranges,
constant runs, BVP inter-beat intervals (0.3--2.0~s), and
wrist-acceleration magnitude. Samples were finite and
detected beat intervals were plausible. Stress-window motion
was within the cohort range (80th percentile), but EDA and
temperature were highly quantized and motion was not minimal.
This is not a validated device-specific quality index, and
residual motion/contact artifact remains an alternative
explanation.
\begin{table}[t]
\centering
\caption{Robustness and selective performance. Covered
metrics condition on windows receiving a prediction.}
\label{tab:results}
\setlength{\tabcolsep}{3.2pt}
\begin{tabular}{lcc}
\toprule
 & \textbf{WESAD} & \textbf{Stress-Predict}\\
\midrule
Subjects & 15 & 35\\
Mean accuracy / F1 & 0.930 / 0.799 & 0.739 / 0.154\\
Pearson $r$ ($p$) & $-0.607$ (.016) & $-0.412$ (.014)\\
Spearman $\rho$ ($p$) & 0.016 (.955) & $-0.300$ (.080)\\
Pearson without S14 & 0.185 (.526) & --\\
FP: model / ICCM & 29 / 27 & 94 / 92\\
FP: random / confidence & 29 / 25 & 87 / 71\\
FP paired $p$ & .157 & .317\\
FN: model / ICCM covered & 71 / 65 & 572 / 542\\
Sensitivity: model / covered & .773 / .781 & .129 / .131\\
Specificity: model / covered & .974 / .975 & .951 / .949\\
Mean abstention / coverage & 0.3\% / 96.8\% & 2.5\% / 94.4\%\\
Subjects $>2$-pp accuracy drop & 0 & 2\\
\bottomrule
\end{tabular}
\end{table}
\section{Experiments}

\subsection{Datasets}
\textbf{WESAD}~\cite{schmidt2018wesad} contains multimodal
Empatica E4 recordings from 15 subjects during baseline,
amusement, meditation, and laboratory stress conditions.
We use BVP, EDA, and TEMP for binary classification
($N=15$).

\textbf{Stress-Predict}~\cite{iqbal2022stress} contains
Empatica E4 recordings from 35 subjects during Stroop and
Interview stress tasks. Hyperventilation segments are
excluded; remaining segments are treated as binary
(baseline vs.\ stress, $N=35$).

\subsection{Experimental Setup}
All experiments use LOSO cross-validation.
ICCM calibrates on each test subject's labeled non-stress
windows. The downstream classifier is a Random Forest
(200 trees) trained on 14 time-domain features from
remaining subjects. We set $\alpha=0.05$.

We report Pearson and Spearman associations, Pearson
correlation without Subject~14, and leave-one-subject-out
influence. At matched coverage, we compare ICCM with random
and confidence abstention. Confusion counts include only
covered windows; abstention is not a correct prediction.
Subject-paired FP changes use a two-sided Wilcoxon signed-rank
test.

\begin{figure}[!t]
    \centering
    \includegraphics[width=\columnwidth]{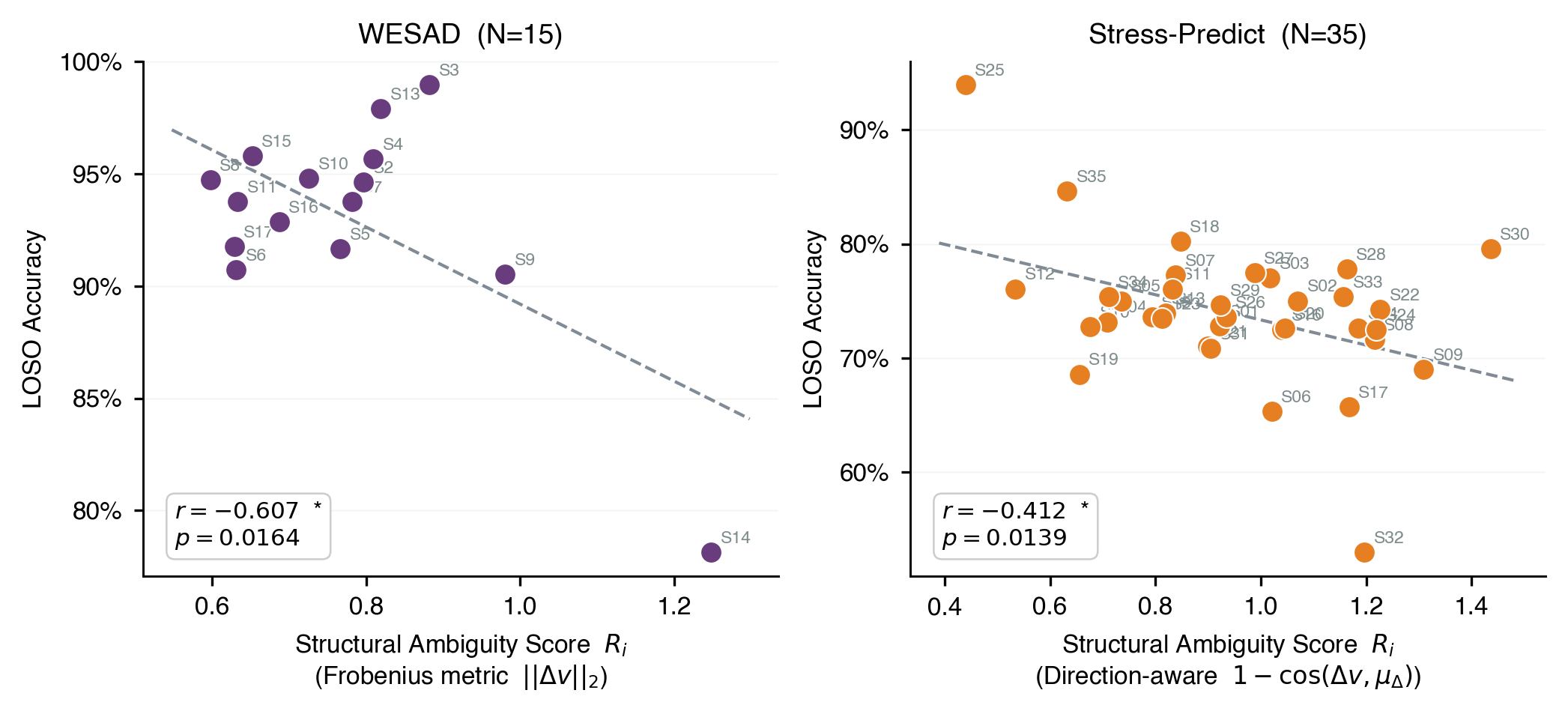}
    \caption{Full-cohort Pearson associations between
    structural ambiguity and LOSO accuracy. WESAD is
    high-leverage: excluding Subject~14 gives $r=0.185$
    ($p=0.526$), and Spearman $\rho=0.016$ ($p=0.955$).
    Stress-Predict Spearman $\rho=-0.300$ ($p=0.080$).}
    \label{fig:moneyplot}
\end{figure}

\subsection{Results}

\textbf{Structural Ambiguity Detection.}
Full-cohort Pearson association is significant in each
dataset (Table~\ref{tab:results}), preserving the main result
that greater coupling divergence accompanies lower
subject-level accuracy. Robustness checks narrow its
interpretation: neither Spearman test is significant, and
removing Subject~14 changes WESAD Pearson $r$ from $-0.607$
to 0.185. In influence analysis, 14 of 15 exclusions retain
$p<0.05$; excluding Subject~14 is the sole exception and
reverses the sign. WESAD is therefore high-leverage, while
Stress-Predict provides a second negative Pearson association
with only suggestive rank evidence.

\textbf{Safety Gate Performance.}
ICCM removes two false alerts in each dataset (29 to 27;
94 to 92). Neither paired change is significant, both
Stress-Predict removals occur for one subject, and random and
confidence baselines remove more Stress-Predict false alerts
at matched coverage. Covered sensitivity changes from 0.129
to 0.131 and specificity from 0.951 to 0.949. Two
Stress-Predict subjects lose more than two percentage points
of covered-window accuracy. ICCM therefore supplies a
distinct physiological routing reason but does not
demonstrate a selective-performance advantage.

For Subject~14, ICCM withholds three of 21 true-stress
windows (two abstentions and one deferral); predictions for
the remaining 18 are all false negatives. ICCM detects part
of the anomalous interval but does not repair the motivating
missed-stress failure.

\section{Discussion}
The central contribution is retained: individualized coupling
divergence exposes a failure that aggregate accuracy conceals
and provides an interpretable signal external to classifier
confidence. The expanded analysis also bounds that
contribution. WESAD is driven by a high-leverage case,
Stress-Predict has low mean F1, confidence thresholding
removes more false alerts, and configuration selection is
post hoc. Overlapping calibration windows preclude a formal
coverage claim, and one Random Forest establishes
classifier-independent implementation rather than
architecture-independent performance.

Reliability across repeated sessions, window-length
sensitivity, device-specific signal-quality indices,
selective-risk curves with cluster-bootstrap uncertainty,
additional classifiers, and naturalistic cohorts remain
future work. ICCM should complement classifier uncertainty
and clinical escalation: ``safe'' denotes a safety-oriented
system objective, not proof that abstention or a non-stress
decision is harmless.
\section{Conclusion}
ICCM preserves the paper's main finding that personalized
coupling divergence can reveal structurally unsupported
inputs and severe individual classifier failure. Across two
datasets, negative Pearson associations motivate this signal,
while robustness and selective-outcome analyses prevent
overinterpretation. ICCM is a transparent candidate component
for safer wearable stress systems, not yet a validated
stand-alone safety mechanism.

\balance

\bibliographystyle{IEEEtran}

\bibliography{references}

\end{document}